%% file: sample-sigconf-authordraft.tex
\documentclass[sigconf, arxiv]{acmart}
\AtBeginDocument{%
  
\pagestyle{fancy}
\fancyhf{} 
\fancyfoot[C]{\thepage}
}

\renewcommand\footnotetextcopyrightpermission[1]{}

\usepackage{caption}
\usepackage{graphicx}
\usepackage{subcaption}
\usepackage{multirow}
\usepackage{booktabs}    

\begin{document}

\title{DUKAE: DUal-level Knowledge Accumulation and Ensemble for Pre-Trained Model-Based Continual Learning}

\author{Songze Li}
\affiliation{%
  \institution{Harbin Institute of Technology}
    \city{Harbin}
    \country{China}
}
\email{lisongze@stu.hit.edu.cn}

\author{Tonghua Su}
\affiliation{%
  \institution{Harbin Institute of Technology}
    \city{Harbin}
    \country{China}
}
\email{thsu@hit.edu.cn}

\author{Xu-Yao Zhang}
\affiliation{%
  \institution{State Key Laboratory of Multimodal Artificial Intelligence Systems, CASIA}
  \institution{School of Artificial Intelligence, UCAS}
    \city{Beijing}
    \country{China}
}
\email{xyz@nlpr.ia.ac.cn}

\author{Qixing Xu}
\affiliation{%
  \institution{Harbin Institute of Technology}
    \city{Harbin}
    \country{China}
}
\email{xuqixing@stu.hit.edu.cn}

\author{Zhongjie Wang}
\affiliation{%
  \institution{Harbin Institute of Technology}
    \city{Harbin}
    \country{China}
}
\email{rainy@hit.edu.cn}

\input{sec/0_abstract}

\maketitle

\input{sec/1_intro}

\input{sec/2_related_work}

\input{sec/3_preliminary}

\input{sec/4_method}

\input{sec/5_experiment}

\input{sec/6_conclusion}

\input{main.bbl}

\end{document}

%% file: sec/0_abstract.tex
\begin{abstract}
Pre-trained model-based continual learning (PTMCL) has garnered growing attention, as it enables more rapid acquisition of new knowledge by leveraging the extensive foundational understanding inherent in pre-trained model (PTM). 
Most existing PTMCL methods use Parameter-Efficient Fine-Tuning (PEFT) to learn new knowledge while consolidating existing memory. 
However, they often face some challenges. 
A major challenge lies in the misalignment of classification heads, as the classification head of each task is trained within a distinct feature space, leading to inconsistent decision boundaries across tasks and, consequently, increased forgetting. 
Another critical limitation stems from the restricted feature-level knowledge accumulation, with feature learning typically restricted to the initial task only, which constrains the model's representation capabilities. 
To address these issues, we propose a method named DUal-level Knowledge Accumulation and Ensemble (DUKAE) that leverages both feature-level and decision-level knowledge accumulation by aligning classification heads into a unified feature space through Gaussian distribution sampling and introducing an adaptive expertise ensemble to fuse knowledge across feature subspaces.
Extensive experiments on CIFAR-100, ImageNet-R, CUB-200, and Cars-196 datasets demonstrate the superior performance of our approach. 
\end{abstract}

%% file: sec/1_intro.tex
\begin{figure}[t]
    \centering
    \includegraphics[width=0.48\textwidth]{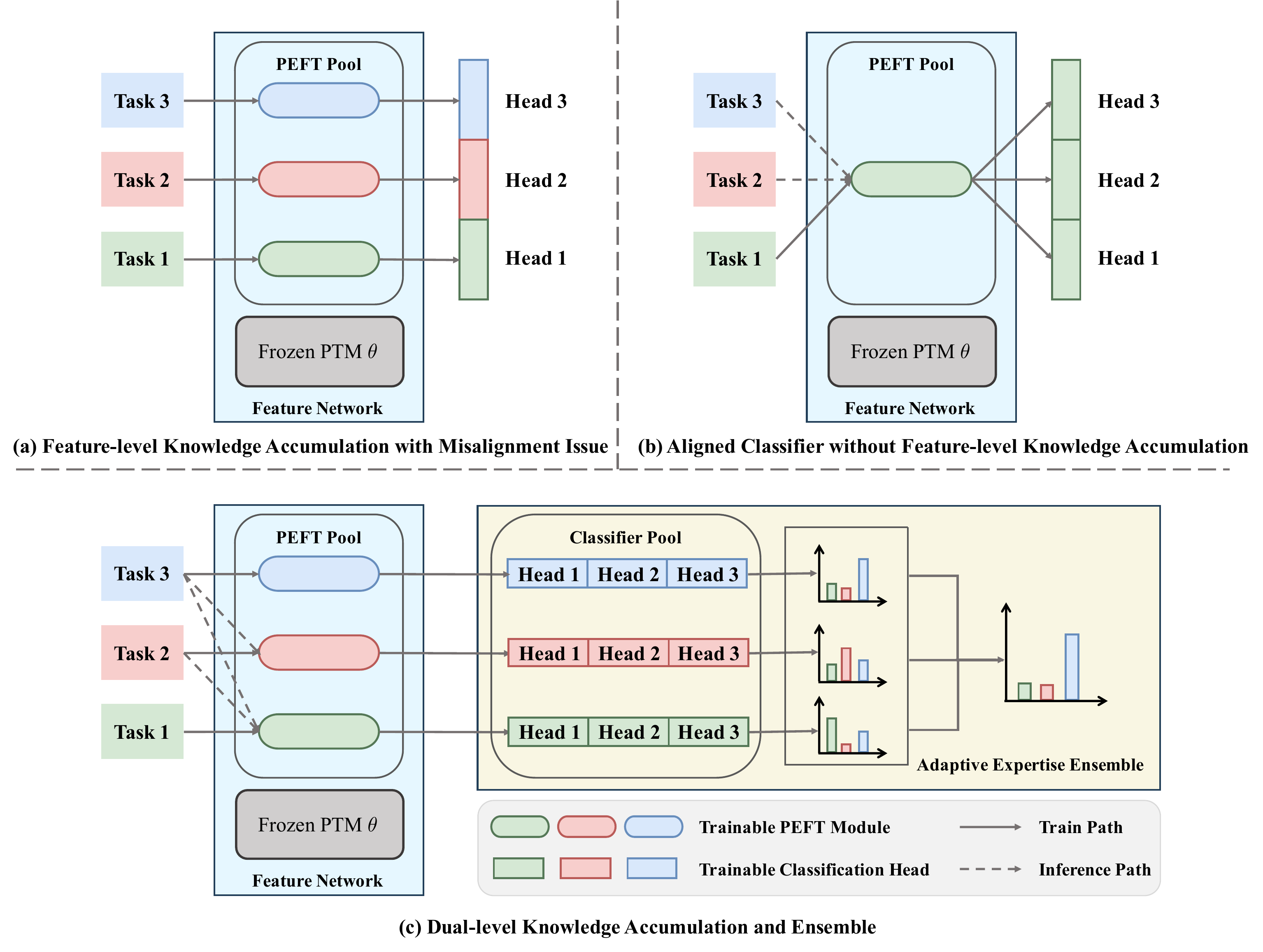}
    \caption{Different PTMCL methods compared with our method. 
    (a) PTMCL methods with misalignment issue. Classification heads are learned in different feature subspaces which are defined by task-specific PEFT modules and then kept fixed. (b) PTMCL methods merely fine-tune feature network with initial task data. Classification heads are learned in same feature space but lack the accumulation of feature knowledge for future tasks. (c) Our method leverages both feature-level and decision-level knowledge accumulation. Each task-specific feature subspace has a corresponding aligned classifier and memory is consolidated through the ensemble of subspace classification results.}
    \label{fig:different_ptm_based_methods}
\end{figure}

\section{Introduction}
\label{sec:intro}

Continual learning seeks to incrementally acquire new knowledge while retaining previously learned information from a continuous data stream.  
Traditionally, continual learning involves model starting from a randomly initialized parameter space and progressively accumulating new knowledge. 
In recent years, with the widespread application of pre-trained model (PTM) in natural language processing (NLP) \cite{devlin2018bert, brown2020language} and computer vision \cite{kirillov2023segment, radford2021learning}, there has been an increasing amount of research in the field of PTM-based continual learning (PTMCL).
Benefiting from the powerful foundational knowledge of PTM, PTMCL methods are akin to standing on the shoulders of giants, where knowledge is accumulated on top of an advanced starting point. 
These approaches significantly outperform traditional continual learning methods, which start from scratch and accumulate knowledge incrementally.

Current PTMCL approaches can generally be categorized into two types. 
The first type \cite{SLCA} treats the PTM's parameters as a new starting point, updating whole PTM parameters continuously without any additional parameters. 
The second type \cite{L2P, DualPrompt, codaprompt, LAE, sprompts}, which is more commonly used, keeps the PTM's parameters fixed and acquires new knowledge through Parameter-Efficient Fine-Tuning (PEFT) \cite{ding2023parameter}. 
These PEFT-based methods fine-tune additional parameters to learn task-specific knowledge, preserving the strong representation capabilities of the PTM. 
Typically, they fine-tune a small portion of parameters for each task, caching these additional parameters to retain knowledge. 
During inference, memory for previous tasks is maintained either by selecting the most relevant PEFT modules corresponding to the inference sample \cite{L2P, DualPrompt, sprompts} or by integrating across all task-specific PEFT modules \cite{codaprompt}.

Despite their advantages, these methods face significant challenges. 
A major issue is that each task-specific classification head is learned within different feature spaces and remains fixed once learned \cite{L2P, DualPrompt, codaprompt, LAE}, leading misalignment of classification heads problem (see figure (a) in Fig. \ref{fig:different_ptm_based_methods}). 
The misalignment of classification heads across tasks results in an inconsistency for inter-task comparison, leading to misclassification and, consequently, increased forgetting.
Some methods \cite{ADAM, RanPAC} avoid the misalignment issue by limiting feature network learning to the initial task only. (see figure (b) in Fig. \ref{fig:different_ptm_based_methods}). 
However, these methods rely solely on the first task's data to train the feature network, without leveraging subsequent data to enhance representation capabilities. 
Consequently, their performance is limited by the feature discrimination ability.

To solve these challenges, we propose a novel DUal-level Knowledge Accumulation and Ensemble (DUKAE) method that pioneeringly leverages both feature-level and decision-level knowledge accumulation by learning task-specific feature subspaces and corresponding subspace-aligned classifiers, followed by an innovative ensemble method to integrate knowledge from these subspaces (see figure (c) in Fig. \ref{fig:different_ptm_based_methods}). 
Typically, we first learn task-specific feature network modules for each new task with PEFT, incorporating self-supervised learning (SSL) to enhance the representation capabilities of the feature network.
The fine-tuned PEFT modules, along with those from previous tasks, are accumulated in a PEFT module pool, with each module defining an independent feature subspace. 
To tackle the misalignment problem, we train aligned classifiers for each feature subspace using Gaussian distribution, which is stored for each category across all existing feature subspaces. 
Finally, with our novel adaptive expertise ensemble, our approach can effectively leverage the accumulated feature-level knowledge from PEFT modules and decision-level knowledge from subspace-aligned classifiers, thereby enhancing memory retention. 
Our contributions are summarized as follows:
\begin{itemize}
\item We propose a pioneering dual-level knowledge accumulation PTMCL approach that leverages both feature-level and decision-level knowledge accumulation.
\item We propose a novel adaptive expertise ensemble to facilitate the efficient ensemble of knowledge from different feature subspaces.
\item We conduct extensive empirical evaluations, demonstrating that our method achieves state-of-the-art (SOTA) performance.
\end{itemize}

%% file: sec/2_related_work.tex
\begin{figure}[t]
    \centering
    \includegraphics[width=0.45\textwidth]{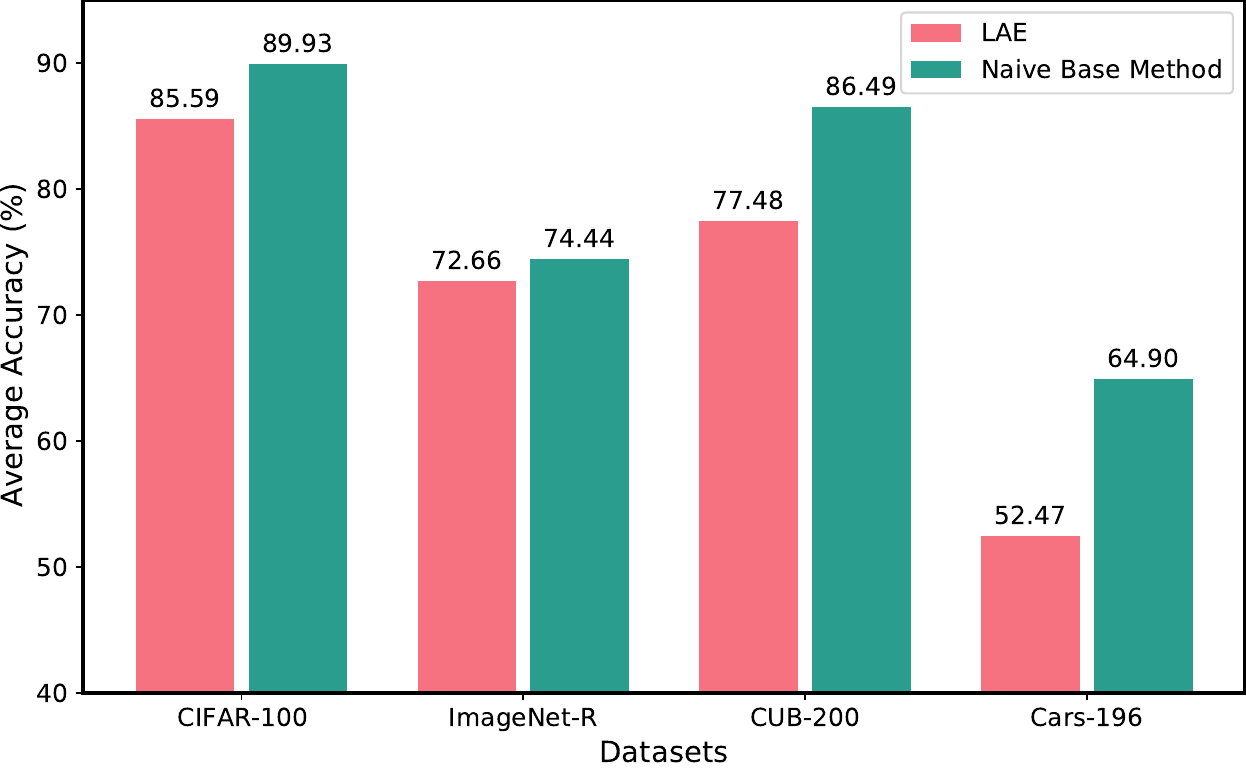}
    \caption{Comparison between our naive base method and LAE method (which suffers from misalignment problem) across four datasets under a 10 tasks continual learning setting. The y-axis represents average accuracy after learning the last task. The comparison is performed under identical network architecture and parameters configurations, illustrating that addressing the misalignment issue effectively mitigates forgetting.
}
    \label{fig:misalignment}
\end{figure}

\section{Related Work}

\textbf{Continual Learning (CL)}, also known as incremental learning or lifelong learning, is a research area focused on enabling models to learn from a continuous data stream without catastrophic forgetting.  
It can be typically categorized into three scenarios: Class-Incremental Learning (CIL), Domain-Incremental Learning (DIL), and Task-Incremental Learning (TIL) \cite{van2019three, de2021continual, CLKD},  in which CIL is the most challenging and widely studied. 
Recent continual learning methods can be broadly classified into three methodological paradigms. 
Replay-based methods mitigate forgetting by storing and replaying past experiences during learning new tasks \cite{rebuffi2017icarl, castro2018end}.
Regularization-based approaches incorporate additional terms in the loss function to protect important parameters of previous tasks, thus maintaining old knowledge \cite{li2017learning, kirkpatrick2017overcoming}. 
Architectural strategies involve dynamically expanding the model or isolating parameters specific to each task to manage new information effectively while retaining prior knowledge \cite{mallya2018packnet, aljundi2017expert}.

\textbf{Parameter-Efficient Fine-Tuning (PEFT)} has emerged as a significant advancement in adapting PTMs to downstream tasks with minimal additional parameters.
Adapter-Tuning \cite{adaptertunning} initially introduces this concept by inserting lightweight, learnable modules into pre-trained transformers.
Prompt-Tuning \cite{prompttuning}  and Prefix-Tuning \cite{prefixtuning} introduce the idea of modifying input prompts or hidden tokens, achieving notable success in NLP tasks.
In the realm of vision transformers, VPT \cite{VPT} and AdapterFormer \cite{adapterformer} extend these ideas to visual tasks.
LoRA \cite{LORA} proposes learning low-rank matrices updates to efficiently fine-tune large models, while SSF \cite{SSF} focuses on scaling and shifting operations within the model for better adaptation.
NOAH \cite{NOAH} leverages a neural architecture search algorithm to design optimal prompt modules for large vision models.

\textbf{Pre-trained model-based continual learning (PTMCL)}  is attracting growing attention  as it enables rapid learning of new knowledge by leveraging the robust foundational knowledge provided by PTM.
L2P \cite{L2P} is the first to introduce PTM into continual learning.
DualPrompt \cite{DualPrompt} uses complementary prompts for task-invariant and task-specific instructions to manage sequential learning.
S-Prompts \cite{sprompts} applies independent prompting across domains with cross-entropy loss for training and K-NN for domain identification.
CODA-Prompt \cite{codaprompt} employs a weighting mechanism to generate prompts, enhancing end-to-end task sequence learning without rehearsal.
ADAM \cite{ADAM} aggregates embeddings from pre-trained and adapted models for classifier construction.
EASE \cite{EASE} utilizes expandable feature subspaces with lightweight adapters, ensuring efficient model updating without conflict.
SLCA \cite{SLCA} integrates slow learning and classifier alignment, improving continual learning by progressively reducing learning rates.
RanPAC \cite{RanPAC} prevents forgetting with training-free random projectors and class-prototype accumulation.

%% file: sec/3_preliminary.tex
\begin{figure}[t]
    \centering
    \begin{subfigure}[t]{0.22\textwidth} 
        \centering
        \includegraphics[width=\textwidth]{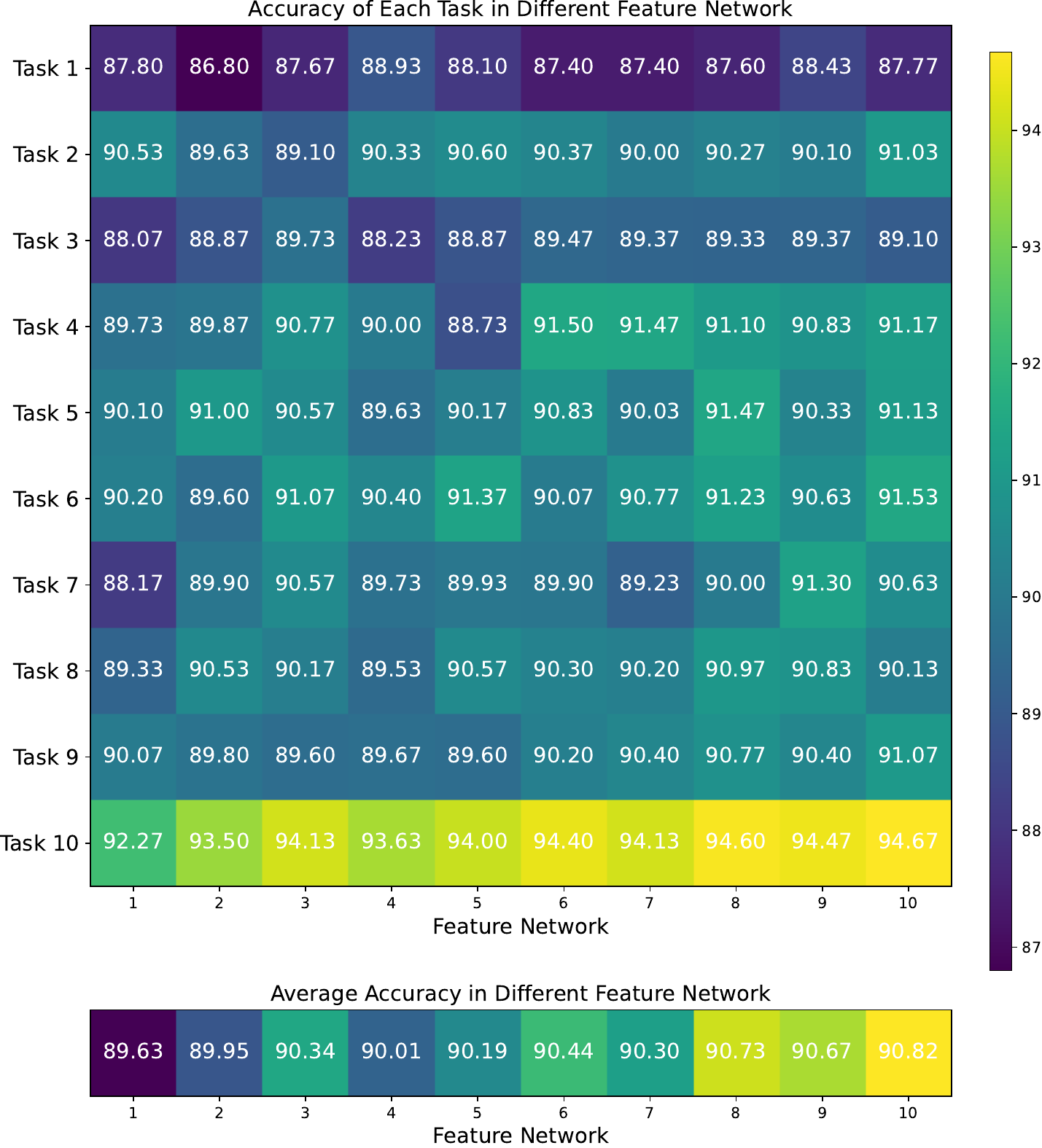}
        \caption{CIFAR-100}
        \label{fig:feature_subspace_cifar100}
    \end{subfigure} %
    \hspace{0.2cm} 
    \begin{subfigure}[t]{0.22\textwidth} 
        \centering
        \includegraphics[width=\textwidth]{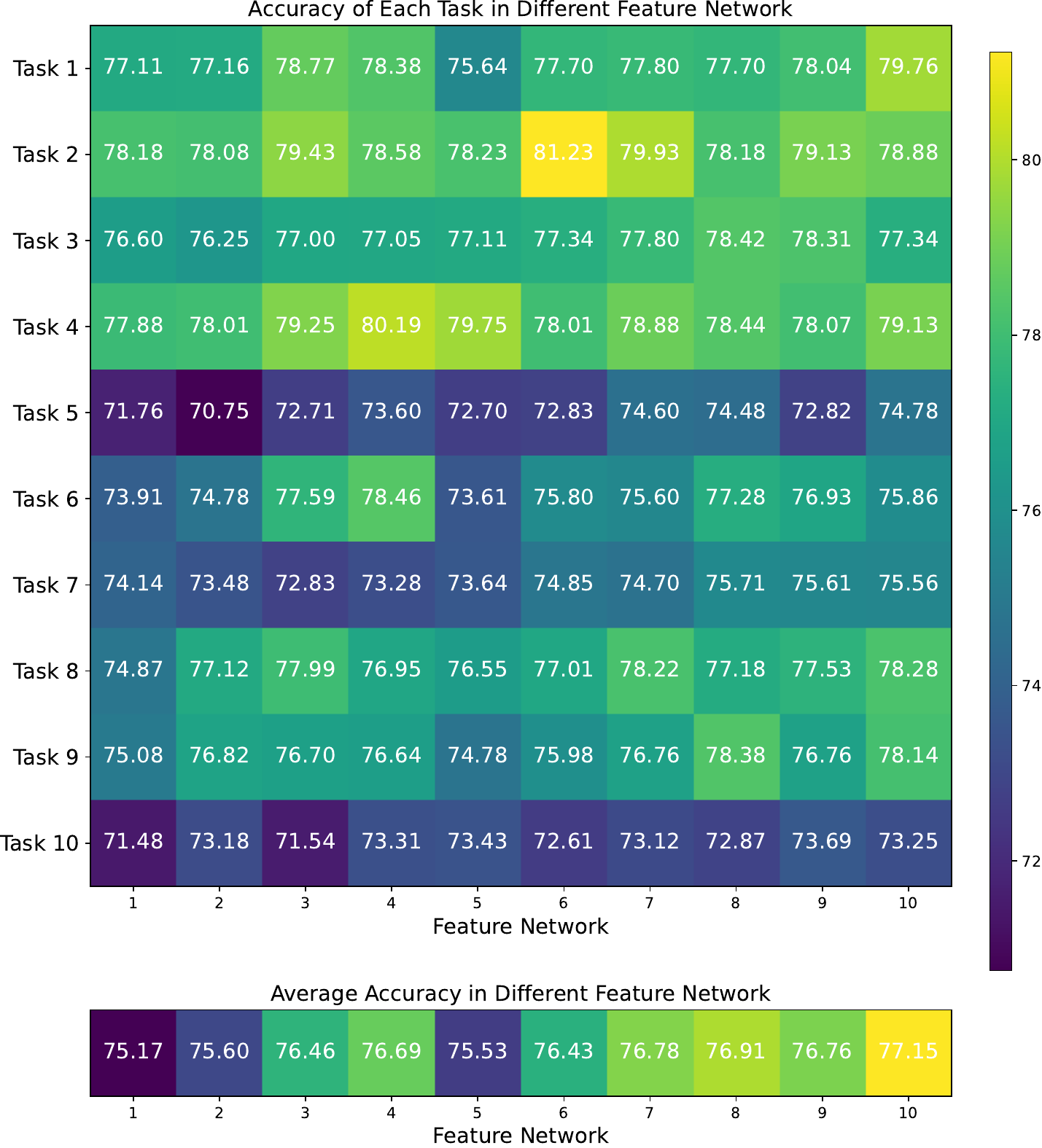}
        \caption{ImageNet-R}
        \label{fig:feature_subspace_imagenet}
    \end{subfigure} %


    \begin{subfigure}[t]{0.22\textwidth} 
        \centering
        \includegraphics[width=\textwidth]{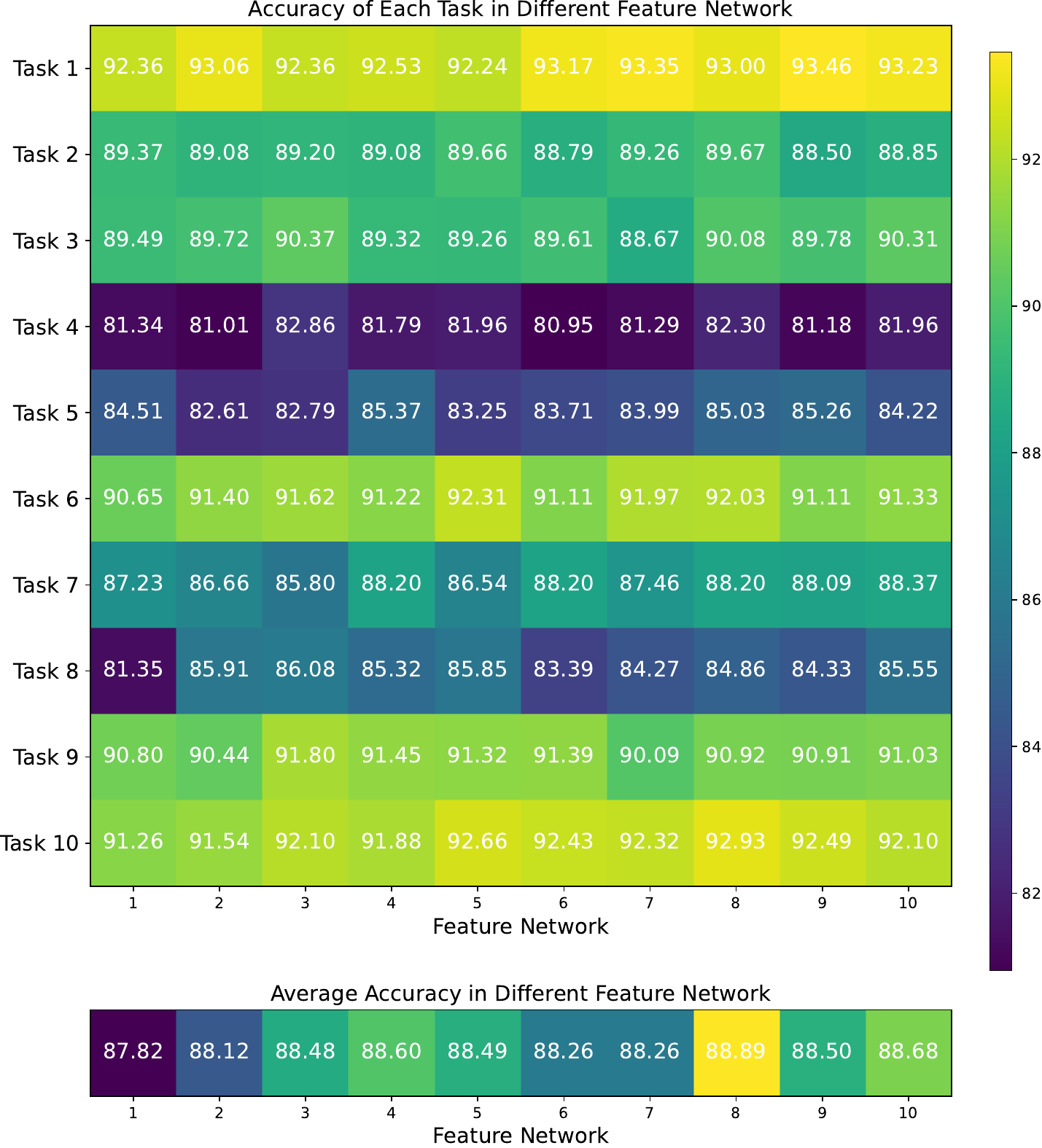}
        \caption{CUB-200}
        \label{fig:feature_subspace_cub200}
    \end{subfigure} %
    \hspace{0.2cm} 
    \begin{subfigure}[t]{0.22\textwidth} 
        \centering
        \includegraphics[width=\textwidth]{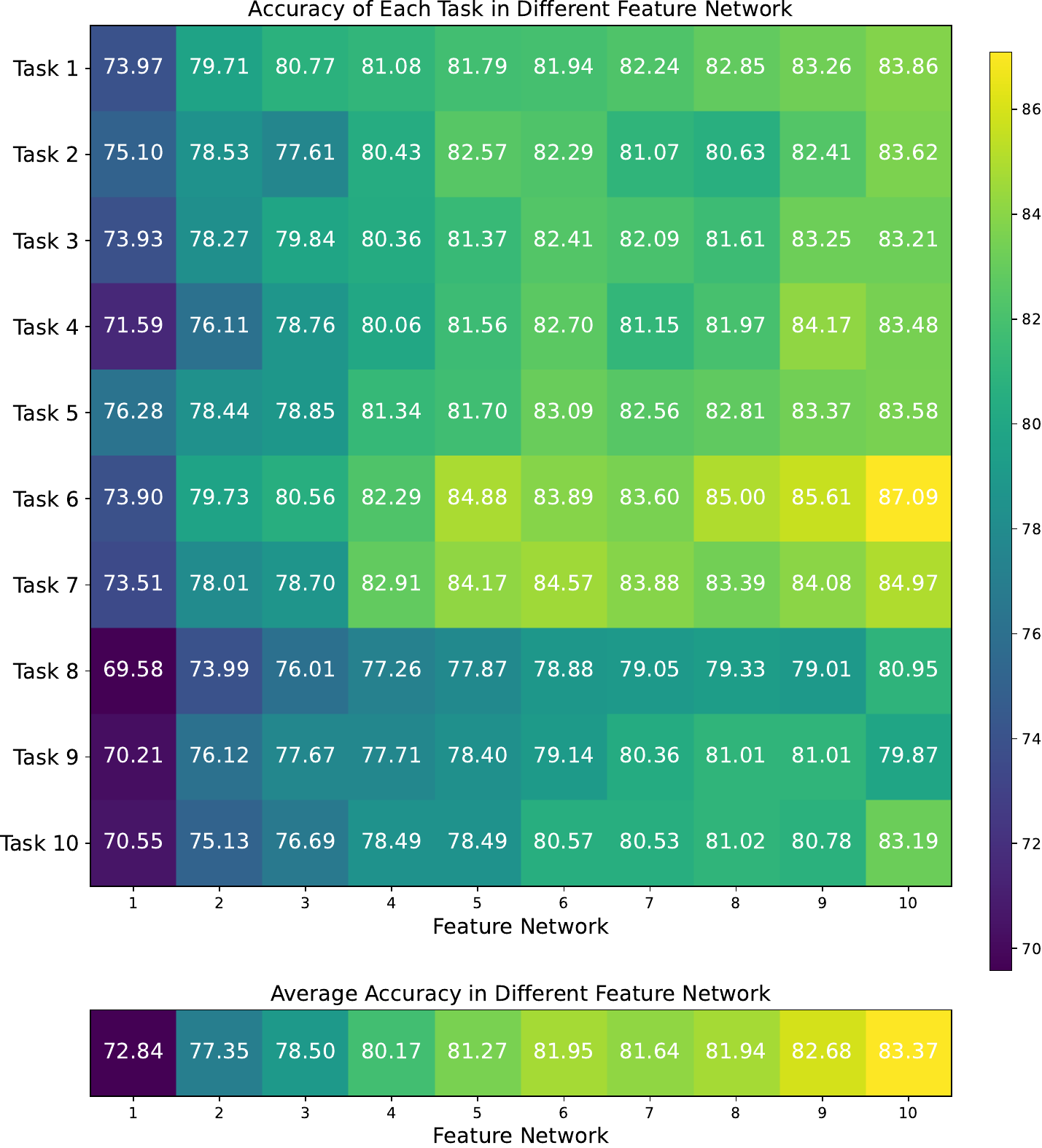}
        \caption{Cars-196}
        \label{fig:feature_subspace_cars196}
    \end{subfigure} %
    
    \caption{Discriminative capability of different feature networks across four datasets. We split each dataset into 10 tasks and train classifiers for all tasks using fixed feature network $f(\cdot; \theta, \mathcal{P}_t)$ learned with dataset of different tasks $t$. The $10\times10$ heatmap represents task-wise classification accuracy, while the $1\times10$ heatmap indicates the average accuracy across tasks for each feature network. The results reveal distinct task-specific discriminative patterns and significant variations in overall feature discrimination capabilities for each feature network.}
    \label{fig:feature_subspace}
\end{figure}

\section{Preliminary}
\subsection{Formulation of PTMCL}

Continual learning aims to progressively acquire new knowledge from a sequential series of tasks data $\{D_1, D_2, \ldots, D_T\}$ while retaining previously learned knowledge.
For each task $t$, the training set $D_t$ is defined as $D_t = \{(x_{t,n}, y_{t,n})\}_{n=1}^{N_t}$, where $N_t$ denotes the number of data-label pairs, $x_{t,n} \in X_t$ represents the input samples, and $y_{t,n} \in Y_t$ are the associated labels. 
Each task $t$ introduces a set of new classes $C_t$, where the number of classes is denoted by $|C_t|$, and there is no overlap in the class sets across different tasks, i.e., $Y_t \cap Y_{t'} = \emptyset$ for $t \ne t'$.

Currently, most PTMCL methods primarily rely on PEFT techniques for new tasks adaptation by fine-tuning a small set of additional parameters while keeping the PTM fixed. 
These techniques enables the model to efficiently learn new task-specific knowledge without compromising the original PTM’s core representation capabilities.
We denote the PTMCL model as $g(f(\cdot; \theta, \mathcal{P}), \varphi)$, where $f(\cdot; \theta, \mathcal{P})$ represents the feature network of the model, with $\theta$ being the parameters of the PTM and $\mathcal{P}$ the additional parameters of the PEFT module (e.g., Adapter \cite{adaptertunning}, LoRA \cite{LORA}). 
The classifier for the downstream tasks is denoted as $g(\cdot; \varphi)$, where $\varphi$ refers to the parameters of classifier.
The PTM module parameters $\theta$ remain fixed during the continual learning, while the PEFT module parameters $\mathcal{P}$ and the classifier parameters $\varphi$ are incrementally fine-tuned as new tasks arrive. 
The mathematical formulation of PTMCL can be generalized as optimizing the following objective function:

\begin{equation}
    \min_{\mathcal{P}, \varphi} \sum_{(x, y) \in D_{1:T}} \mathcal{L}\left(g\left(f\left(x; \theta, \mathcal{P}\right); \varphi\right), y\right) + \mathcal{L}_{\text{reg}}(\mathcal{P}),
\end{equation}
where $\mathcal{L}$ represents the loss function measures the discrepancy between the predicted and true labels, while $\mathcal{L}_{\text{reg}}$ serves as a regularization term to control the complexity of the PEFT module.

\begin{figure*}[t]
    \centering
    \includegraphics[width=1\textwidth]{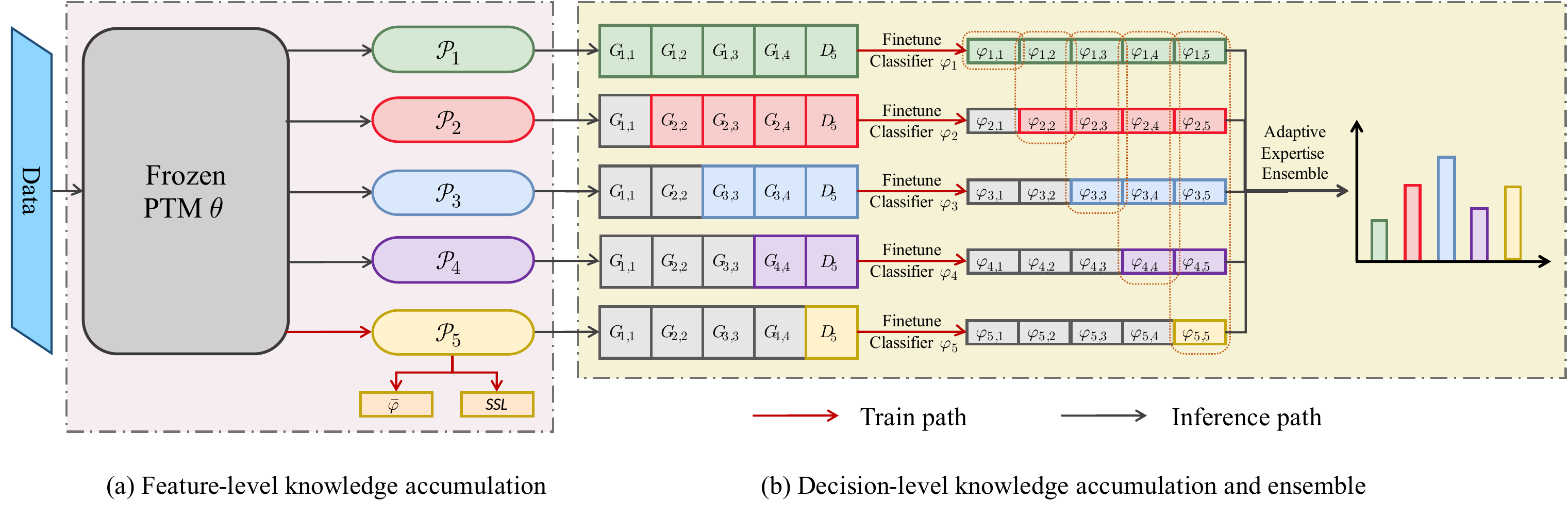}
    \caption{Illustration of our method, exemplified through the learning process on Task 5. (a) Feature-level knowledge accumulation process. The auxiliary classification head \( \bar{\varphi} \) and the SSL branch are employed to facilitate the learning of the PEFT module \( \mathcal{P}_5 \), which is then cached for cumulative learning. (b) Decision-level knowledge accumulation and ensemble process. Using current task data, feature representations of \( D_5 \) are extracted in each feature subspace. These are then combined with Gaussian distributions \( G_{i,j} \) from prior tasks $j$ to train classifiers \( \varphi_i \) specific to each feature subspace $i$. Finally, adaptive expertise ensemble is applied to the ensemble of final output.
}
    \label{fig:whole_framework}
\end{figure*}

\subsection{Naive Base Method}
Previous PTMCL methods primarily employ PEFT to accumulate task-specific feature-level knowledge. 
However, many methods \cite{L2P, DualPrompt, codaprompt, LAE} often neglect the importance of aligning classification heads across tasks. 
Typically, once a task-specific classification head is trained, it remains fixed, leading to different tasks' classification heads learned in different feature spaces. 
This misalignment of classification heads creates inconsistent decision boundaries across tasks, making it difficult to effectively compare classification results, consequently increasing  forgetting.

To address the misalignment problem, we propose a naive base method that maintains all classification heads of each task in a unified feature space. 
The key insight is to continuously refine the classification head using features from all encountered classes in the same feature space. 
Initially, we train a PEFT module $\mathcal{P}_1$ on the first task's data $D_1$ atop the PTM $\theta$. 
The resulting feature network $f(\cdot; \theta, \mathcal{P}_1)$ is used to train a classifier $g(\cdot; \varphi_1)$ for task 1. 
For each subsequent task $t$, we employ features from all encountered classes to refine the classification heads, maintaining their alignment within feature space $\mathcal{P}_1$. 
Specifically, for classes in current task, their features can be directly extracted through the feature network $f(\cdot; \theta, \mathcal{P}_1)$. 
Meanwhile, for previously encountered classes, we utilize their stored Gaussian distributions $\mathcal{G}_{1,c} = \mathcal{N}(\mu_{1,c}, \sigma_{1,c}^2)$, which are computed and cached at the end of each task for each class $c$ in feature space $\mathcal{P}_1$, to sample representative features. 
This process can be formulated as:

\begin{equation}
\begin{split}
    \min_{\varphi_{1}} &\left[ \sum_{(x, y) \in D_t} \mathcal{L}\left(g\left(f(x; \theta, \mathcal{P}_1); \varphi_{1}\right), y\right) \right. \\
    &\left. + \sum_{c=1}^{|C_{1:t-1}|} \sum_{(x_c, y_c) \sim \mathcal{G}_{1,c}} \mathcal{L}\left(g(x_c; \varphi_{1}), y_c\right) \right].
\end{split}
\end{equation}

Here, $\mathcal{L}$ is the loss function measuring the discrepancy between predicted and true labels, which usually is cross entropy loss. $|C_{1:t-1}|$ denotes the number of old classes.

To empirically verify the relationship between misalignment and catastrophic forgetting, we conducted a controlled comparative study between our naive base method and the LAE \cite{LAE} approach (which suffers from misalignment) while maintaining identical network architecture and parameters configurations. 
The experimental results reveal that mitigating misalignment effectively alleviates forgetting and enhances overall performance (see Fig. \ref{fig:misalignment}).

%% file: sec/4_method.tex
\section{DUal-level Knowledge Accumulation and Ensemble}

\begin{table*}[t]
\centering
\caption{Result of LAA and IAA for baseline methods and our approach across four datasets. * means results from our re-implementation.}
\label{tab:all_result}
\resizebox{\textwidth}{!}{%
\begin{tabular}{l|cc|cc|cc|cc}
\specialrule{1pt}{0pt}{0pt}
\multirow{2}{*}{Method} & \multicolumn{2}{c|}{CIFAR-100} & \multicolumn{2}{c|}{ImageNet-R} & \multicolumn{2}{c|}{CUB-200} & \multicolumn{2}{c}{Cars-196} \\  
                        & LAA (\%)       & IAA (\%)      & LAA (\%)       & IAA (\%)       & LAA (\%)      & IAA (\%)     & LAA (\%)      & IAA (\%)     \\ \hline
L2P \cite{L2P}                      & 82.76±1.17                        & 88.48±0.83                    & 66.49±0.40                    & 72.83±0.56                    & 62.21±1.92                    & 73.83±1.67                    & 38.18±2.33                    & 51.79±4.19          \\
DualPrompt \cite{DualPrompt}        & 85.56±0.33                        & 90.33±0.33                    & 68.50±0.52                    & 72.59±0.24                    & 66.00±0.57                    & 77.92±0.50                    & 40.14±2.36                    & 56.74±1.78          \\
CODA-Prompt \cite{codaprompt}       & 86.56±0.77                        & 90.61±0.36                    & 75.25±0.56                    & 81.26±0.76                    & 72.63±0.76                    & 80.54±0.54                    & 44.89±0.61                    & 58.91±0.37          \\
LAE \cite{LAE}                      & 85.59±0.46                        & 89.96±0.44                    & 72.66±0.63                    & 78.91±0.89                    & 77.48±0.94                    & 85.83±0.68                    & 52.47±1.46                    & 64.08±1.01          \\
ADAM$^*$ \cite{ADAM}                & 87.46±0.03                        & 92.20±0.06                    & 66.70±0.21                    & 75.18±0.07                    & 86.77±0.02                    & 91.32±0.01                    & 41.77±7.73                    & 53.78±7.76          \\
EASE$^*$ \cite{EASE}                & 87.73±0.17                        & 92.29±0.19                    & 75.89±0.28                    & 81.67±0.20                    & 86.62±0.08                    & 91.20±0.03                    & 37.58±0.25                    & 51.00±0.14          \\
SLCA \cite{SLCA}                    & 91.53±0.28                        & 94.09±0.87                    & 77.00±0.33                    & 81.17±0.64                    & 84.71±0.40                    & 90.94±0.68                    & 67.73±0.85                    & 76.93±1.21          \\
SLCA++ \cite{SLCAPP}                & 91.69±0.15                        & 94.47±0.72                    & \underline{79.78±0.16}        & \underline{84.31±0.73}        & 86.59±0.29                    & 91.63±0.72                    & 73.97±0.22                    & 79.46±0.80          \\
RanPAC$^*$ \cite{RanPAC}            & \underline{92.10±0.17}            & \underline{95.13±0.06}        & 77.44±0.34                    & 83.06±0.30                    & \underline{89.09±0.23}        & \underline{92.86±0.03}        & \underline{74.69±0.92}        & \underline{82.24±0.41}          \\ \hline
Ours w/ Adapter                     & 92.25±0.15                        & 95.17±0.09                    & 81.07±0.28                    & 86.02±0.38                    & 89.18±0.12                    & 92.86±0.09                    & \textbf{84.71±0.47}           & \textbf{88.29±0.46} \\
Ours w/ LoRa                        & \textbf{92.39±0.13}               & \textbf{95.21±0.07}           & \textbf{81.42±0.30}           & \textbf{86.03±0.38}           & \textbf{89.39±0.06}           & \textbf{93.01±0.15}           & 84.23±0.25                    & 87.71±0.22          \\ \specialrule{1pt}{0pt}{0pt}
\end{tabular}%
}
\end{table*}

Although our naive base method addresses the classifier misalignment issue, its performance is inherently constrained by the limited feature discrimination ability of the feature network  $f(\cdot; \theta, \mathcal{P}_1)$. 
Since $f(\cdot; \theta, \mathcal{P}_1)$ primarily learns features related to the first task, it does not accumulate new feature knowledge for subsequent tasks. 
However, feature networks trained on different tasks' data may exhibit varying discriminative capabilities. 
As demonstrated in Fig. \ref{fig:feature_subspace}, when we train classifiers for all tasks using fixed feature network $f(\cdot; \theta, \mathcal{P}_t)$ learned with dataset of different task $t$, each network demonstrates distinct discriminative power. 
This observation naturally leads to the idea of combining multiple feature networks to enhance overall feature discrimination ability. 
Therefore, we propose a dual-level knowledge accumulation and ensemble approach, which aggregates and leverages both feature-level and decision-level knowledge.
Fig. \ref{fig:whole_framework} illustrates our proposed method.

\subsection{Feature-level Knowledge Accumulation}
To achieve the feature-level knowledge accumulation, we use an expansion strategy same as  \cite{EASE} to store new task-specific PEFT modules. 
Typically, for each subsequent task $t$, a new PEFT module $\mathcal{P}_t$ is trained using the current task's data $D_t$, and we cache the task-specific PEFT modules for all seen tasks $\{\mathcal{P}_1, \ldots, \mathcal{P}_t\}$. 
For each task \(t\), we train the associated PEFT module \(\mathcal{P}_t\) using a cross-entropy loss function:
\begin{equation}
    \min_{\mathcal{P}_t, \bar{\varphi}} \sum_{(x, y) \in D_t} \mathcal{L}_{CE}\left(g(f(x; \theta, \mathcal{P}_t); \bar{\varphi}), y\right),
\end{equation}
where $g(\cdot;\bar{\varphi})$ is a temporary auxiliary classifier used for training PEFT module which only predict current task.

To further boost feature discrimination ability, we integrate an auxiliary self-supervised learning (SSL) branch \cite{lee2020self, PASS}. 
This branch simply classify four new classes by rotating input data by 0, 90, 180, and 270 degrees. 
The SSL loss is defined as:
\begin{equation}
    \mathcal{L}_{SSL} = \mathcal{L}_{CE}\left(g(f(Rot(x); \theta, \mathcal{P}_t); \bar{\varphi}_{SSL}), Rot(y)\right), 
\end{equation}
where \(Rot(x)\) denotes the rotation operation on input \(x\), $Rot(y)$ denotes corresponding four classes label after rotation. $g(\cdot;\bar{\varphi}_{SSL})$ is auxiliary classifier for SSL loss.

The final loss function for training each PEFT module $\mathcal{P}_t$ is formulated as:
\begin{equation}
    \mathcal{L}_{\mathcal{P}_t} = \mathcal{L}_{CE}\left(g(f(x; \theta, \mathcal{P}_t); \bar{\varphi}), y\right) + \alpha \cdot \mathcal{L}_{SSL},
\end{equation}
where \(\alpha\) is a hyperparameter controlling the contribution of the SSL branch.

\subsection{Decision-level Knowledge Accumulation}
To accumulate the decision-level knowledge, we propose training a unified classifier for the feature subspace defined by each PEFT module $\mathcal{P}_t$. 
We aim to use the same approach as our naive base method, generating class features using Gaussian distributions to train classifiers. 
However, when learning the $t$-th task, we cannot access the data of previous tasks. 
This means we can only obtain the Gaussian distributions of current task data in all trained feature subspaces, but not the Gaussian distributions of previous task data in the current feature subspace.

To approximate the feature distributions of old class $\mathcal{G}_{t,c}$ in current feature subspace $\mathcal{P}_t$, we employ a method that utilizes the Gaussian distribution $\mathcal{G}_{T_c,c}$, computed in the feature subspace $\mathcal{P}_{T_c}$, where $T_c$ is the task to which class $c$ belongs. 
Once the feature subspace $\mathcal{P}_t$ corresponding to the $t$-th task is learned, the classifiers  $g(\cdot; \varphi_k)$ (where $k \leq t$) associated with all previously accumulated feature subspaces $\mathcal{P}_k$  need to be fine-tuned to enable recognition of the new task. 
The objective function for fine-tuning the $k$-th classifier can be formulated as:
\begin{equation}
\begin{split}
    \min_{\varphi_{k}} &\left[ \sum_{(x, y) \in D_t} \mathcal{L}\left(g(f(x; \theta, \mathcal{P}_k); \varphi_k), y\right) \right. \\
    &\left. + \sum_{c=1}^{|C_{1:t-1}|} \sum_{(x_c, y_c) \sim \mathcal{G}_{k,c}} \mathcal{L}\left(g(x_c; \varphi_k), y_c\right) \right],
\end{split}
\label{eq:classifier}
\end{equation}
where $\mathcal{G}_{k,c}$ represents the Gaussian distributions of old class $c$ in the feature subspace $\mathcal{P}_k$.
After fine-tuning all subspace-specific classifiers, we calculate and store the Gaussian distribution information of task $t$ data in all feature subspaces $\mathcal{P}_1$ to $\mathcal{P}_t$ .

\subsection{Adaptive Expertise Ensemble} \label{sec:aee
}

Having achieved knowledge accumulation at both feature and decision levels, we proceed to synthesize this complementary knowledge through ensemble. 
However, in the process of refining classifiers corresponding to different feature subspaces, only the classifier associated with the first feature subspace $\mathcal{P}_1$ employs real Gaussian distribution for all encountered classes. 
For the feature subspaces corresponding to later tasks, as we lack access to old task data, the Gaussian distributions corresponding to previous task classes are approximated rather than derived from actual data distributions in the feature subspace.
This approximation introduces systematic errors in the classification results for previous tasks within the classifiers of these subsequent feature subspaces.
To avoid these errors, we propose an adaptive expertise ensemble strategy that fuses classifier outputs based on their expertise, excluding outputs from classifiers on tasks they are not specialize in.

Typically, let \( Z_k = g(f(x; \theta, \mathcal{P}_k); \varphi_k) \), where \( Z_{k,t} \) represents the sub prediction for task \( t \) in \( Z_k \) , $\varphi_{k,t}$ represents the sub classification head in $\varphi_k$ specific to task $t$, and \( S_t(x) \) is the ensemble result for task \( t \). 
The ensemble process is given by the following formula:
\begin{equation}
    S_t(x) = \frac{1}{t} \sum_{k=1}^{t} Z_{k,t}.
\end{equation}

The final prediction result is obtained by concatenating the ensemble outputs of all \( T \) tasks and then taking the maximum value:
\begin{equation}
    \hat{y} = \arg \max_y \left( \text{Concat}\left[S_1(x), S_2(x), \dots, S_T(x)\right] \right).
\end{equation}

%% file: sec/5_experiment.tex
\begin{figure}[t]
    \centering
    \begin{subfigure}[t]{0.22\textwidth}  
        \centering
        \includegraphics[width=\textwidth]{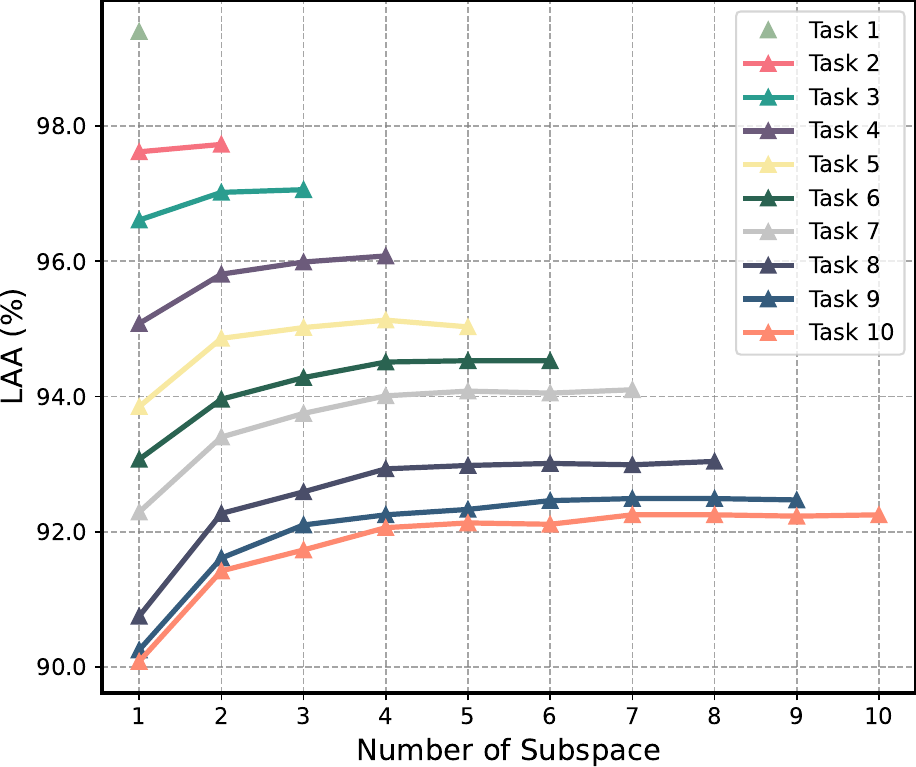}  
        \caption{CIFAR-100}
        \label{fig:ablation_cifar100}
    \end{subfigure}
    \begin{subfigure}[t]{0.22\textwidth}  
        \centering
        \includegraphics[width=\textwidth]{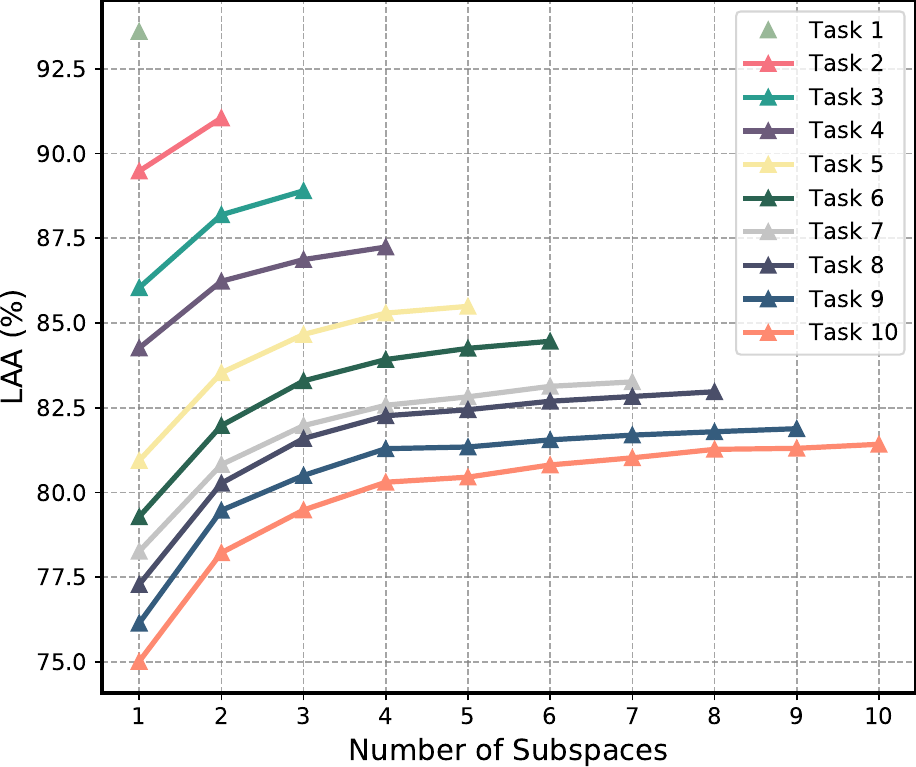}  
        \caption{ImageNet-R}
        \label{fig:ablation_imagenet_r}
    \end{subfigure}


    \begin{subfigure}[t]{0.22\textwidth}  
        \centering
        \includegraphics[width=\textwidth]{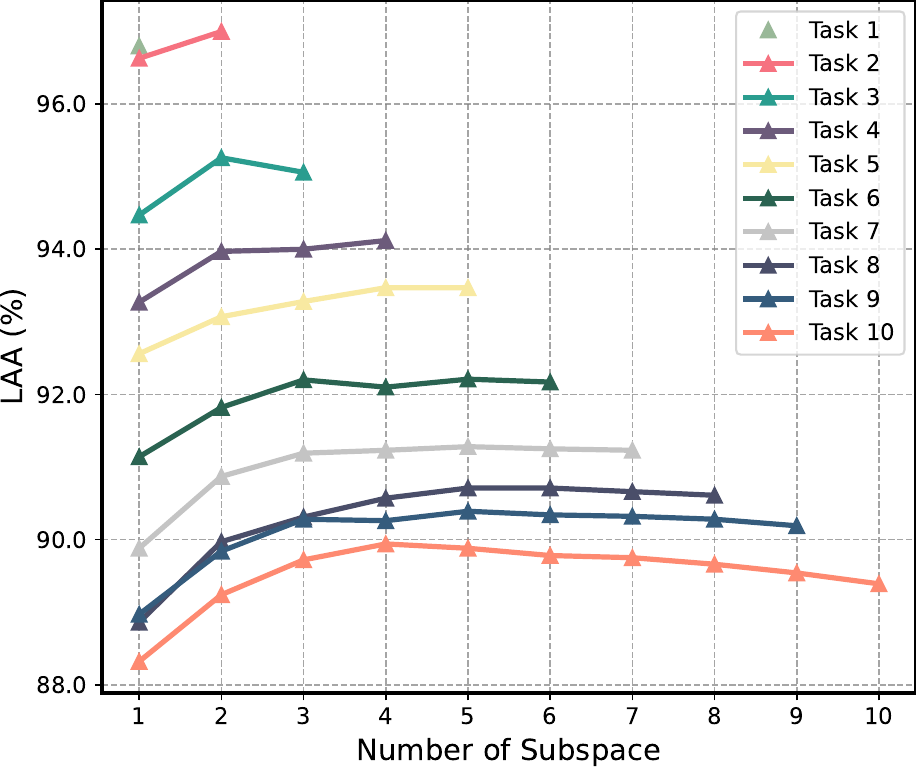}  
        \caption{CUB-200}
        \label{fig:ablation_cub200}
    \end{subfigure}
    \begin{subfigure}[t]{0.22\textwidth}  
        \centering
        \includegraphics[width=\textwidth]{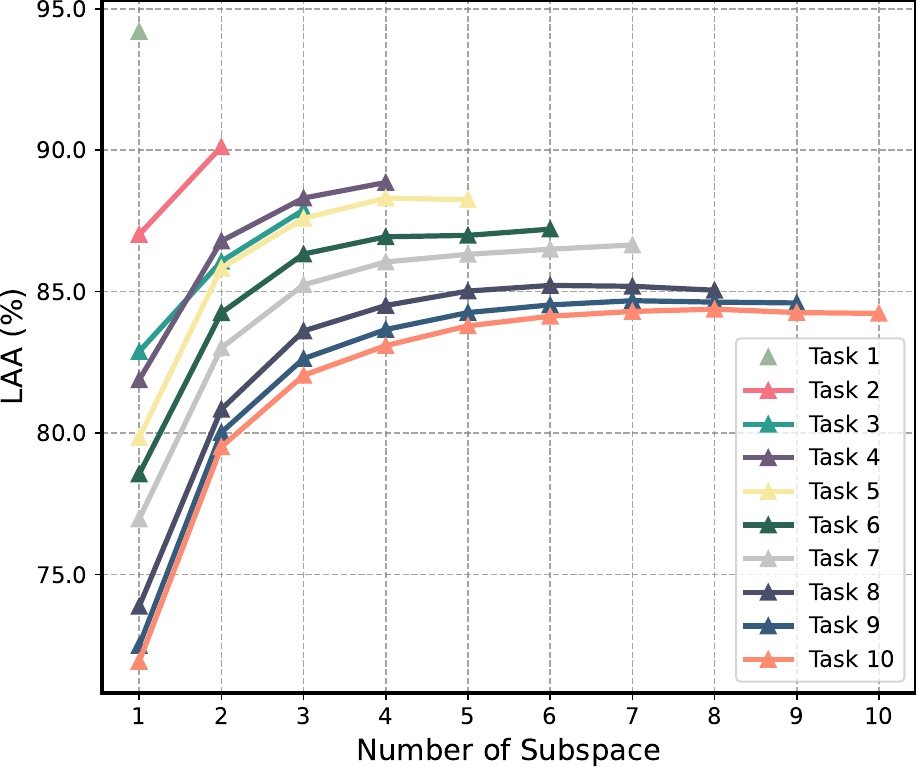}  
        \caption{Cars-196}
        \label{fig:ablation_cars196}
    \end{subfigure}
    
    \caption{Performance comparison of ensemble with varying number of subspaces. Each figure shows LAA results from the first task to the tenth task using varying number of subspaces for ensemble across four datasets. Each line in the figure represents the performance of the ensemble with different number of subspaces after learning the corresponding task. For the first task, a maximum of one subspace can be utilized for ensemble, and this increases incrementally such that the tenth task allows for the use of up to ten subspaces for ensemble.}
    \label{fig:ablation_num_of_classifiers}
\end{figure}

\section{Experiments}

In this section, we describe the experimental setups and present the results of our proposed method compared with various PTMCL methods.

\subsection{Benchmark and Evaluation Metrics}

We follow SLCA \cite{SLCA} to evaluate our method on four widely-used benchmarks—CIFAR-100 \cite{krizhevsky2009learning}, ImageNet-R \cite{hendrycks2021many}, CUB-200 \cite{wah2011caltech}, and Cars-196 \cite{krause20133d}—where the first two datasets focus on relatively coarse-grained classification tasks, while the latter two emphasize fine-grained classification.
CIFAR-100 consists of 100 classes of small-scale images across 100 classes, with 50,000 training images and 10,000 test images. 
ImageNet-R comprises 200 classes of large-scale images, with 24,000 train images and 6,000 test images styled in various artistic renditions.
CUB-200 focuses on fine-grained bird classification which contains 200 bird species and split into 5,994 training images and 5,794 test images.
Cars-196 dataset includes 16,185 images of 196  fine-grained vehicles which is divided into 8,144 training images and 8,041 test images.
In our approach, all datasets are divided into 10 tasks to simulate the continual learning scenario. 
Each task in CIFAR-100 comprises 10 classes, while in ImageNet-R and CUB-200, each task contains 20 classes. 
For Cars-196, the first task consists of 16 classes, with each subsequent task including 20 classes.

Regarding the evaluation metrics, we follow \cite{SLCA, SLCAPP} to present the average accuracy of all classes after learning the last task, which we refer to as \textbf{L}ast task \textbf{A}verage \textbf{A}ccuracy (LAA, corresponding to ``Last-Acc" in \cite{SLCA, SLCAPP}). We also calculate the average accuracy of all previously seen classes after learning each incremental task, which we denote as \textbf{I}ncremental tasks \textbf{A}verage \textbf{A}ccuracy (IAA, corresponding to ``Inc-Acc" in \cite{SLCA, SLCAPP}).

\subsection{Baselines}

We compare our proposed method against several PTMCL approaches, including L2P \cite{L2P}, DualPrompt \cite{DualPrompt}, CODA-Prompt \cite{codaprompt}, LAE \cite{LAE}, ADAM \cite{ADAM}, EASE \cite{EASE}, SCLA \cite{SLCA}, SLCA++ \cite{SLCAPP} and RanPAC \cite{RanPAC}. 
Due to differences in result presentation and specific experimental settings, we reproduced the results of the ADAM, EASE, and RanPAC methods using the authors' official open-source code, ensuring the use of the same PTM parameters and maintaining the same class order in the datasets for a fair comparison. 
For the remaining methods, their experimental results are primarily referenced from SLCA++ \cite{SLCAPP}. 
To ensure a fair comparison, we follow \cite{ADAM} to apply a random seed 1993 to shuffle the class order of the dataset before dividing it into 10 tasks.

\begin{table}[t]
\centering
\caption{LAA results with and without SSL loss ($\mathcal{L}_{\text{SSL}}$) applied to learn PEFT modules across four datasets.}
\label{tab:ablation_ssl_loss}
\resizebox{0.48\textwidth}{!}{%
\begin{tabular}{l|cc|cc}
\specialrule{1pt}{0pt}{0pt}
\multirow{2}{*}{Dataset} & \multicolumn{2}{c|}{Adapter} & \multicolumn{2}{c}{LoRA} \\

                         & w/o SSL & w/ SSL  & w/o SSL & w/ SSL  \\ \hline
CIFAR-100                & 92.07   & 92.25 ($\uparrow$0.18) & 92.12   & 92.39 ($\uparrow$0.27) \\
ImageNet-R               & 80.44   & 81.07 ($\uparrow$0.63) & 80.62   & 81.42 ($\uparrow$0.80) \\
CUB-200                  & 89.14   & 89.18 ($\uparrow$0.04) & 89.33   & 89.39 ($\uparrow$0.06) \\
Cars-196                 & 84.62   & 84.71 ($\uparrow$0.09) & 83.99   & 84.23 ($\uparrow$0.24) \\
\specialrule{1pt}{0pt}{0pt}
\end{tabular}}
\end{table}

\subsection{Training Details}

Our experiments are conducted based on the implementation framework provided by LAE, with necessary modifications to incorporate our proposed methodology. 
We use a ViT-B/16 model as the backbone, employing the IN21K-Sup PTM parameters, which is trained on the ImageNet-21K dataset with supervised learning.
In our approach we utilize Adapter, LoRA as PEFT modules, injecting these modules into all the layers of the ViT-B/16 model. 

For the training of PEFT modules, all datasets employ the Adam optimizer with a learning rate of 0.0005 and a batch size of 64. 
Specifically, training is conducted for 10 epochs on CIFAR-100 and CUB-200 , while ImageNet-R and Cars-196 require 50 epochs. 
For classifier fine-tuning, the same configuration is applied across all datasets: classifiers are trained with the SGD optimizer at a fixed learning rate of 0.1 for 30 epochs, with a batch size of 64.
Regarding the $\mathcal{L}_{\text{SSL}}$ loss coefficient used in the training of PEFT module, we conducted a hyperparameter search for each dataset. 
Based on our hyperparameter search, we used 0.05 for CIFAR-100 and Cars-196, 0.01 for CUB-200, and 0.3 for ImageNet-R as the loss coefficient $\mathcal{L}_{\text{SSL}}$. 
The model is trained with NVIDIA GeForce RTX 3090 under the PyTorch framework. 
Each experiment is conducted over three independent trials, using random seeds 1993, 1994, and 1995.

\subsection{Results Analysis}

Table \ref{tab:all_result} presents the performance comparison between our approach and all baseline methods across the CIFAR-100, ImageNet-R, CUB-200, and Cars-196 datasets. 
As shown in the table, our method achieves the best results across all datasets, both in terms of LAA and IAA. Notably, on the ImageNet-R dataset, our method outperforms the best-performing baseline by 1.64\% in LAA and 1.72\% in IAA. For the Cars-196 dataset, the improvements are even more significant, with gains of 10.02\% in LAA and 6.05\% in IAA.
For CIFAR-100 and CUB-200, our method also demonstrates consistent improvements. Specifically, on CIFAR-100, our method achieves an increase of 0.29\% in LAA and 0.08\% in IAA. Similarly, on CUB-200, the improvements are 0.3\% in LAA and 0.15\% in IAA.
Furthermore, we observe that our method performs better with LoRA on CIFAR-100, ImageNet-R, and CUB-200 datasets, while Adapter yields superior performance on the Cars-196 dataset.

\subsection{Ablation Study}
\label{sec:ablation}

\textbf{About the Number of Ensemble Subspaces.} We first conducted an ablation study on the performance impact of the number of ensemble subspaces. 
The Fig. \ref{fig:ablation_num_of_classifiers} illustrates the LAA result when using varying number of subspaces for ensemble across four datasets, from the first task to the tenth task. 
The experiments reveal that, in most cases, more ensemble subspaces do lead to better results, except for the tenth task in the CUB-200 dataset, where performance increases with the number of ensemble subspaces up to a point, after which it experiences a slight decrease. 
For all datasets, performance significantly increases when using two subspaces for ensemble, with subsequent performance gains becoming progressively more gradual and even exhibiting slight fluctuations. 
For example, in the CIFAR-100 dataset, the performance improvement tends to level off once the number of ensemble subspaces exceeds three. 
Among all datasets, Cars-196 and ImageNet-R experience the most significant performance gains due to ensemble. 

\begin{table}[t]
\centering
\caption{LAA results with no-ensemble (NoE), simple ensemble (SE) and our adapter expertise ensemble (AEE).}
\label{tab:ablation_ensemble_strategy}
\begin{tabular}{l|ccc|ccc}
\specialrule{1pt}{0pt}{0pt}
\multirow{2}{*}{Dataset} & \multicolumn{3}{c|}{Adapter} & \multicolumn{3}{c}{LoRA} \\

                         & NoE & SE & AEE  & NoE & SE & AEE  \\ \hline
CIFAR-100                & 90.08   & 86.06   & 92.25  & 90.29  & 87.34   & 92.39  \\
ImageNet-R               & 74.87   & 73.79   & 81.07  & 75.01  & 75.81   & 81.42  \\
CUB-200                  & 87.95   & 79.39   & 89.18  & 88.32  & 81.36   & 89.39  \\
Cars-196                 & 71.16   & 64.31   & 84.71  & 71.92  & 64.72   & 84.23  \\
\specialrule{1pt}{0pt}{0pt}
\end{tabular}
\end{table}

\textbf{About PEFT Module Learning. }
We conducted an ablation study to assess the impact of SSL loss ($\mathcal{L}_{\text{SSL}}$) on the representation capacity of PEFT modules. 
We analyzed the LAA results with and without SSL loss applied to the learning process of PEFT modules across four datasets, as shown in Table \ref{tab:ablation_ssl_loss}. 
As illustrated in the table, SSL effectively enhances the model's representation capability of PEFT modules, promoting feature-level knowledge accumulation and consequently improving the model's resistance to catastrophic forgetting. 
For the two coarse-grained datasets, the SSL loss demonstrates a more pronounced enhancement on the representational power of PEFT modules. 
Notably, on the ImageNet-R dataset, performance improvements of 0.63\% and 0.8\% are achieved using Adapter and LoRA configurations, respectively. 
In comparison, the SSL loss exhibits relatively limited efficacy on the two fine-grained datasets, particularly for the CUB-200 dataset.
For the Cars-196 dataset, the SSL loss shows a more substantial improvement when LoRA is used.

\textbf{About Ensemble Strategy. }

To validate that our proposed adaptive expertise ensemble can effectively integrate useful information from different subspaces, we compared it with a simple ensemble method, which directly applies averaging of the classification results from all subspaces, and also with the strategy in which no ensemble is used (only using the result of subspace corresponding to the first task). The results are shown in Table \ref{tab:ablation_ensemble_strategy}. As can be seen in the table, the simple ensemble method cannot effectively integrate the knowledge from each feature subspace, and its performance is significantly lower than that of our proposed adaptive expertise ensemble. In fact, in most cases, its performance is even worse than the no-ensemble approach, which indicates that the simple ensemble method fails to properly utilize the knowledge from each subspace and may even be detrimental to performance. This further provides evidence that our proposed adaptive expertise ensemble can integrate and utilize knowledge from each subspace correctly and efficiently.

\begin{table}[t]
    \centering
    \caption{Results with different continual learning tasks. The first four dataset results (top block) are obtained using Adapter, while the latter four (bottom block) correspond to LoRA.}
    \label{tab:ablation_cl_tasks}
        \resizebox{0.48\textwidth}{!}{%
        \begin{tabular}{l|cc|cc}
        \specialrule{1pt}{0pt}{0pt}
        \multirow{2}{*}{Dataset} & \multicolumn{2}{c|}{5 Tasks} & \multicolumn{2}{c}{20 Tasks} \\
                                 & LAA (\%) & IAA (\%)  & LAA (\%) & IAA (\%)  \\ \hline
        CIFAR-100       & 92.27±0.13 & 94.67±0.13 & 91.87±0.24 & 95.05±0.05 \\
        ImageNet-R      & 81.64±0.30 & 85.83±0.18 & 80.28±0.08 & 85.86±0.13 \\
        CUB-200         & 89.30±0.09 & 92.61±0.03 & 89.39±0.19 & 93.24±0.14 \\
        Cars-196        & 85.15±0.12 & 88.46±0.09 & 83.75±0.17 & 87.65±0.25 \\ \hline \hline
        CIFAR-100       & 92.49±0.09 & 94.85±0.11 & 92.16±0.16 & 95.10±0.04 \\
        ImageNet-R      & 81.22±0.45 & 85.45±0.07 & 80.52±0.10 & 85.83±0.25 \\
        CUB-200         & 89.65±0.06 & 92.70±0.01 & 89.48±0.18 & 93.31±0.05 \\
        Cars-196        & 84.83±0.13 & 87.77±0.18 & 83.73±0.26 & 86.27±0.98 \\
        \specialrule{1pt}{0pt}{0pt}
        \end{tabular}}
\end{table}

\textbf{About Different Number of CL Tasks. }
To validate the robustness of our proposed method under varying numbers of continual learning tasks, we conducted experiments on four datasets under two settings: 5-task and 20-task. For the Cars-196 dataset, which contains 196 classes, we adapted the task division strategy from the 10-task setup. Specifically, in the 5-task setting, the first task learned 36 classes, while in the 20-task setting, the initial task covered 6 classes, with subsequent classes uniformly distributed across remaining tasks. For other datasets, classes were evenly divided per task. As shown in Table \ref{tab:ablation_cl_tasks}, the results demonstrate that our method achieves consistent performance across both task configurations.
The 20-task setting generally presents greater challenges, as evidenced by lower LAA metrics compared to the 5-task scenario in most cases, except on the CUB-200 dataset when using Adapter, where performance remains stable. 
Despite the increased task complexity, performance degradation in the 20-task scenario is relatively mild and all datasets maintain strong overall performance. 
These results collectively demonstrate the method’s robust adaptability across diverse continual learning scenarios.

\begin{figure*}[t]
    \centering
    
    \begin{subfigure}{0.24\textwidth}  
        \centering
        \includegraphics[width=\textwidth]{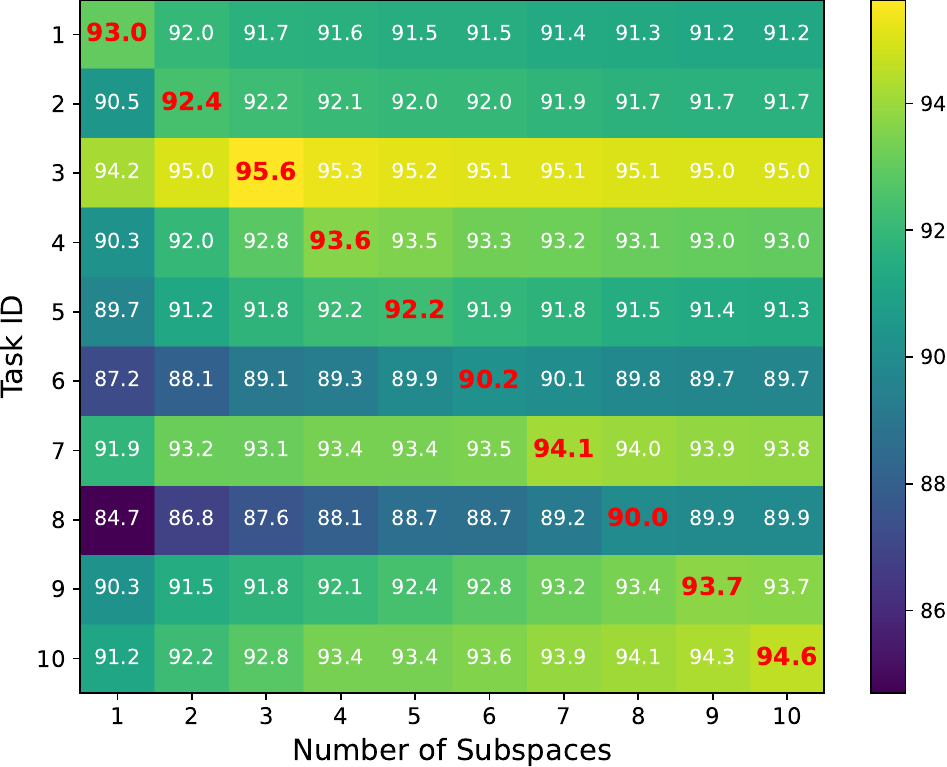}  
        \caption{CIFAR-100}
        \label{fig:task_spe_cifar100}
    \end{subfigure}
    \begin{subfigure}{0.24\textwidth}  
        \centering
        \includegraphics[width=\textwidth]{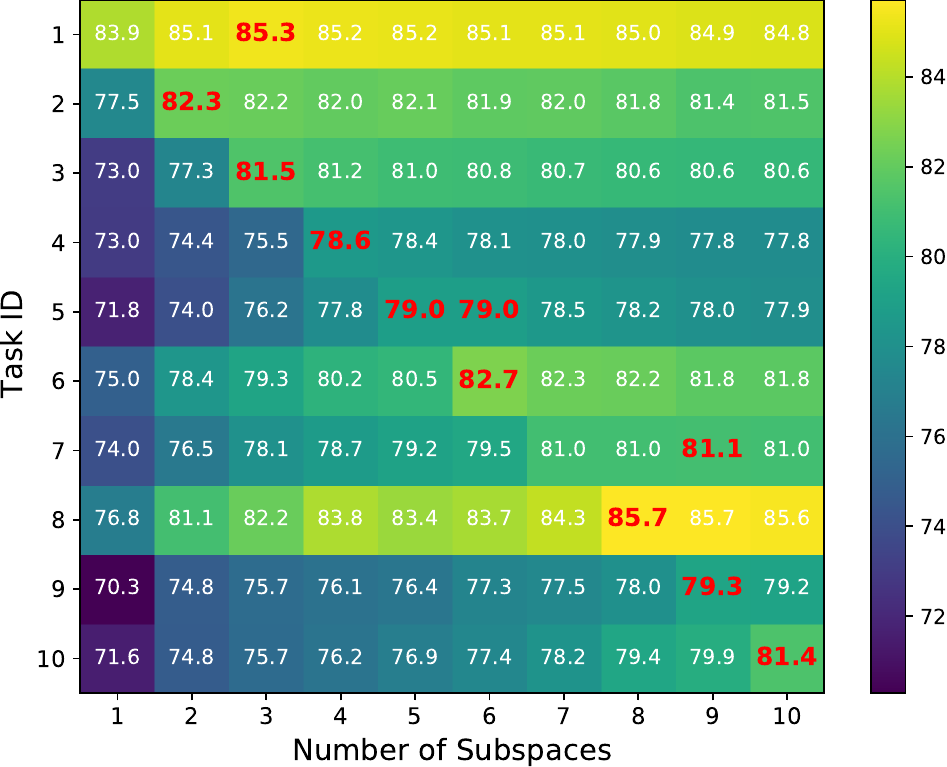}  
        \caption{ImageNet-R}
        \label{fig:task_spe_imagenet_r}
    \end{subfigure}
    \begin{subfigure}{0.24\textwidth}  
        \centering
        \includegraphics[width=\textwidth]{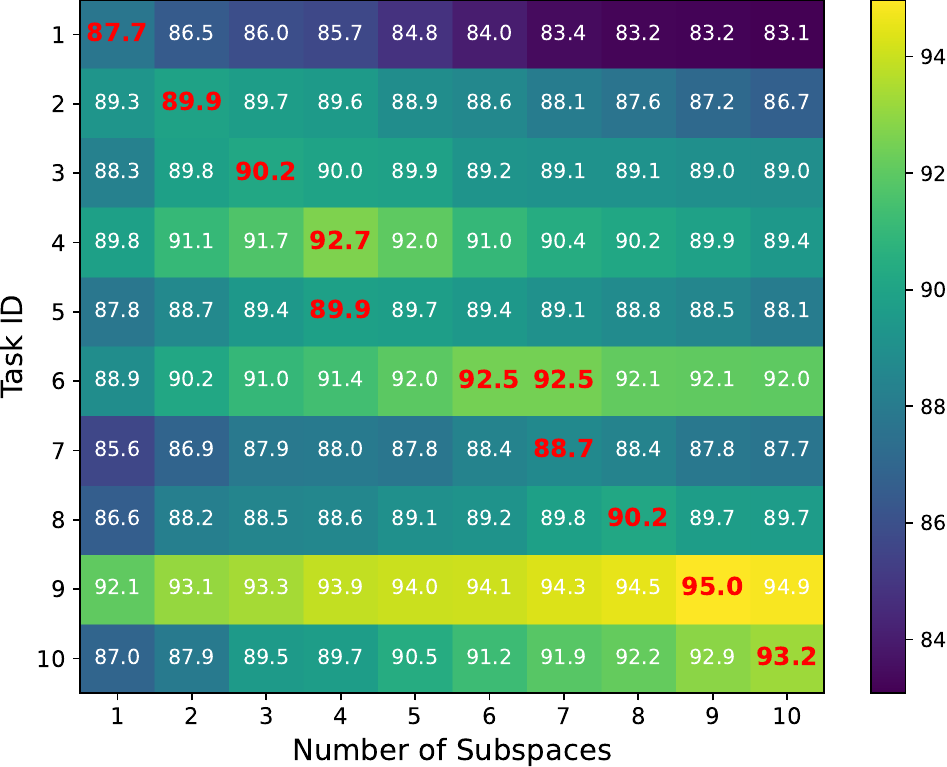}  
        \caption{CUB-200}
        \label{fig:task_spe_cub200}
    \end{subfigure}
    \begin{subfigure}{0.24\textwidth}  
        \centering
        \includegraphics[width=\textwidth]{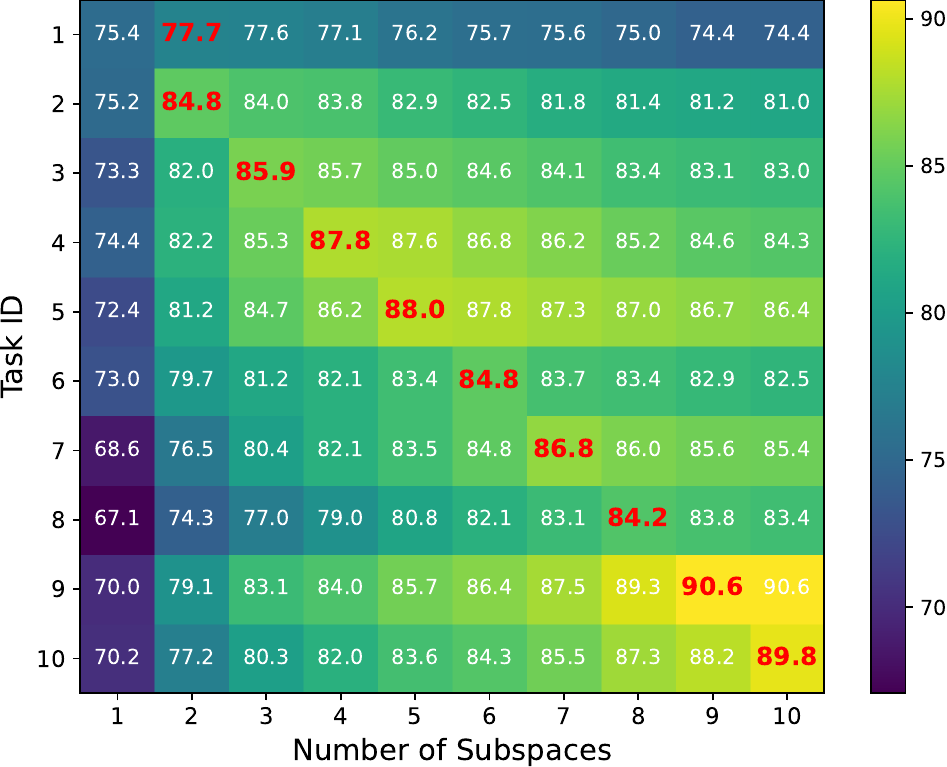}  
        \caption{Cars-196}
        \label{fig:task_spe_cars196}
    \end{subfigure}
    
    \caption{Adaptive expertise ensemble analysis. Each figure shows accuracy of each task when performing ensemble with varying number of subspaces after completing the 10th task. Each row in the figure represents the accuracy of each task when using an ensemble with varying number of subspaces.}
    \label{fig:ensemble_analysis}
\end{figure*}
\subsection{Adaptive Expertise Ensemble Analysis}

To thoroughly investigate the mechanisms of the adaptive expertise ensemble and provide empirical evidence for its efficacy, we systematically analyzed the accuracy for each task when performing an ensemble with varying number of subspaces after completing the 10th task, as illustrated in Fig. \ref{fig:ensemble_analysis}.
Each row in the heatmap represents the change in accuracy for each task as the number of ensemble subspaces increases. 
From the figure, it is evident that, in most cases, the accuracy of each task tends to improve as more subspaces with expertise classifier—trained using the real Gaussian distribution of categories for this task from the corresponding feature subspace, rather than an approximated Gaussian distribution—are incorporated into the ensemble. 
However, when subspaces with non-expertise classifier are introduced into the ensemble, this can lead to classification confusion, resulting in a decrease in the accuracy of the corresponding task.
For example, the classifiers from the first six feature subspaces demonstrate expertise in Task 6, while the latter four exhibit limited proficiency. 
Consequently, the classification accuracy shows a consistent improvement during the ensemble of the first six subspaces but experiences a gradual decline with the ensemble of subsequent subspaces.
Therefore, for the first task, its performance gradually degrades as the number of ensemble subspaces increases, whereas the last task exhibits progressive performance enhancement with subspace aggregation. 
This phenomenon confirms that the adaptive expertise ensemble achieves effective knowledge fusion across subspaces, establishing an optimal performance equilibrium between sequentially learned tasks.
Although specific exceptions exist—such as Task 1 in CUB-200, where excessive ensemble of non-expertise classifiers leads to substantial performance degradation in the specific task that impacts overall accuracy, thereby explaining the accuracy decline observed in Fig. \ref{fig:ablation_cub200} when aggregating more subspaces during Task 10's final phase-our adaptive expertise ensemble effectively capitalizes on domain-specific expertise within each subspace while mitigating interference from irrelevant classifiers, thereby conclusively validating its efficacy.

%% file: sec/6_conclusion.tex
\section{Conclusion and Discussion}

In this paper, we introduced a novel PTMCL approach by accumulating knowledge from both feature-level and decision-level. 
Our method aligns classifiers within consistent feature spaces, and by employing the proposed adaptive expertise ensemble, we facilitate the efficient integration of knowledge from different subspace.
Extensive experiments on four datasets show that our method outperforms SOTA PTMCL methods, significantly improving accuracy and reducing forgetting. 
However, our method still has some limitations. As the number of tasks increases, the storage requirements for Gaussian distributions of each category across different feature subspaces grow quadratically, which raises concerns about offline storage overhead. 
Future work should focus on optimizing this aspect to reduce storage costs. 
Beyond the advancement and limitation, our method can be treated as a multi-agent fusion based continual learning paradigm, which naturally aligns with federated continual learning systems. 
Through subspace coordination, it enables secure cross-device knowledge transfer while preserving localized expertise, particularly promising for smart city deployments that require collaborative model evolution across heterogeneous domains.